\documentclass[sigconf]{aamas}

\usepackage{balance}   
\usepackage{amsmath}
\usepackage{dsfont}
\usepackage{algorithm}
\usepackage{algpseudocode}

\makeatletter

\setcopyright{none}
\renewcommand\footnotetextcopyrightpermission[1]{}
\newcommand{\projecturl}{\url{https://zhanghutony.github.io/ibmarl/}}

\copyrightyear{2027}
\acmYear{2027}
\acmDOI{}
\acmISBN{}

\acmSubmissionID{3029}

\title[Imitation Bootstrapped Multi-Agent RL]{Imitation Bootstrapped Multi-Agent Reinforcement Learning}

\author{Tony H. Zhang}
\authornote{Authors contributed equally to this work.}
\affiliation{
  \institution{University of Utah}
  \city{SLC, UT}
  \country{USA}}
\email{tony.h.zhang@utah.edu}

\author{Connor Mattson}
\authornotemark[1]
\affiliation{
  \institution{University of Utah}
  \city{SLC, UT}
  \country{USA}}
\email{c.mattson@utah.edu}

\author{Daniel S. Brown}
\affiliation{
  \institution{University of Utah}
  \city{SLC, UT}
  \country{USA}}
\email{daniel.s.brown@utah.edu}

\newcommand{\il}{\text{IL}}
\newcommand{\rl}{\text{RL}}
\newcommand{\obs}{\mathbf{o}}
\newcommand{\act}{\mathbf{a}}
\newcommand{\teacher}{\mu_{\psi}}
\newcommand{\cand}{\mathcal{A}}

\begin{abstract}
Demonstrations are an effective way to accelerate single-agent reinforcement learning (RL), but extending this idea to multi-agent reinforcement learning (MARL) is non-trivial as it is difficult and often impossible for a single person to teleoperate multiple agents simultaneously. Round-Robin Behavior Cloning (R2BC) addresses this by letting a single demonstrator control one agent at a time while the others execute their current policies. However, the resulting imitation policies are limited by the quality of the demonstrations. We propose Imitation Bootstrapped Multi-Agent Reinforcement Learning (IBMARL), a framework that extends Imitation Bootstrapped Reinforcement Learning (IBRL) to cooperative teams and uses R2BC to train the imitation policies. At each time step, every agent receives one action proposal from its imitation policy and one from its RL policy, and an ensemble of centralized critics selects among the per-agent combinations of the two. A gated imitation penalty, a soft Q-filtered behavior cloning loss, regularizes each RL policy toward its imitation policy only where the critics value the imitation action more, so that after training each agent executes its RL policy alone, without the centralized critics. On four cooperative multi-agent RL benchmarks, IBMARL surpasses its imitation policies $2.6$--$4.2\times$ faster than demonstration-based baselines on two tasks, is the only method to surpass it on the other two, and exceeds the demonstrator itself on two of the four. With real human demonstrations IBMARL still surpasses imitation learning alone, and does so more than twice as fast as the best baseline. Code, rollout videos and full results are available at \projecturl.
\end{abstract}


\begin{document}

\maketitle
\thispagestyle{plain}
\pagestyle{plain}
\section{Introduction}
\label{sec:intro}
Cooperative multi-agent tasks are difficult for Reinforcement Learning (RL) to solve: the joint-action space grows exponentially with the number of agents, credit assignment across agents is ambiguous, and rewards are often inherently sparse~\cite{albrecht2024marl}. Even with a known model, finding an optimal joint policy for a finite-horizon Decentralized Partially Observable Markov Decision Process (Dec-POMDP) is NEXP-complete~\cite{bernstein2002complexity}, so a learner that must also discover the reward through random joint exploration faces a search that grows with every agent added. In single-agent RL, demonstrations have been shown to help with this exploration problem, as examples of good behavior can often lead agents toward a reward signal even in sparse-reward tasks. Methods such as DQfD~\cite{hester2018dqfd}, DDPGfD~\cite{vecerik2017ddpgfd}, and DDPG + BC~\cite{nair2018overcoming} seed the replay buffer or regularize the learner toward expert data, and Imitation Bootstrapped Reinforcement Learning (IBRL)~\cite{hu2023ibrl} goes further by keeping a behavior-cloned policy in the loop, using it to propose actions during exploration and to improve the bootstrap target (Sec.~\ref{sec:ibrl}).

\begin{figure}[t]
\centering
\includegraphics[width=\columnwidth]{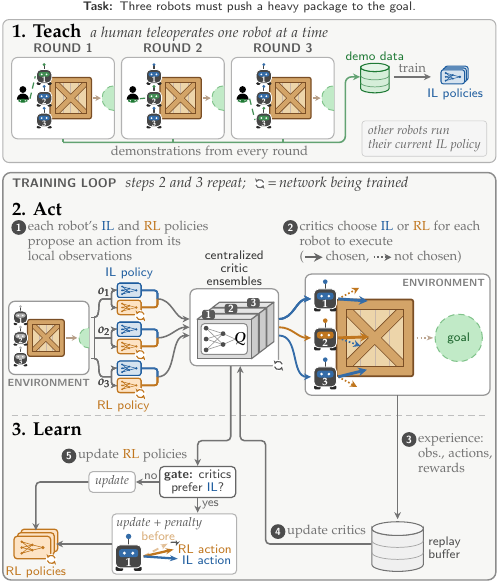}
\caption{IBMARL overview. \normalfont{\textbf{(1) Teach:} a human teleoperates one robot at a time while the others run their current IL policies (R2BC), producing a frozen IL policy for each robot. \textbf{(2) Act:} each robot's IL and RL policies propose an action, and the centralized critic ensembles choose, per robot, which one the team executes. \textbf{(3) Learn:} the experience trains the critics, and each RL policy is pulled toward its IL action only where the critics prefer it. At deployment, each robot runs its RL policy alone.}}
\label{fig:overview}
\end{figure}

We address two obstacles that block the extension of such frameworks to MARL: \textbf{(i)}~\emph{Demonstrations.} Nearly all multi-agent imitation learning (MAIL) assumes joint demonstrations, in which an expert must simultaneously command all $N$ agents---data a lone human operator cannot realistically provide for collaborating teams~\cite{mattson2026r2bc,turco2024cognitive}. To address this, we leverage Round-Robin Behavior Cloning (R2BC)~\cite{mattson2026r2bc} (Sec.~\ref{sec:r2bc}), which has the demonstrator cycle through controlling each agent of the team, while the other agents execute their current learned policies, yielding decentralized cooperative policies from realistic single-human demonstrations. However, R2BC inherits the limitations of behavior cloning: covariate shift, a performance ceiling at the demonstrator, and a prohibitive demonstration burden for the last increments of task performance. \textbf{(ii)}~\emph{Decentralized execution.} IBRL's deployed policy relies on the critic at every step to choose between the IL action and the RL action. For a team, a critic that judges joint actions must have access to every agent's observation and action, so keeping it during deployment would require centralized execution or communication. To avoid this, the RL actors must be deployable on their own, but nothing in IBRL ensures that they are, and the RL policies lag behind the policy that was actually executed during training (Sec.~\ref{sec:ablations}).

We address both obstacles with \textbf{Imitation Bootstrapped Multi-Agent Reinforcement Learning (IBMARL)}, which keeps the R2BC policy in the loop rather than only in the data (Fig.~\ref{fig:overview}). Assuming access to a task reward, we ask: \emph{Can MARL be bootstrapped from policies trained from realistic single-agent demonstrations, and can the result be deployed without the critics?} During training (Fig.~\ref{fig:overview}, Act), every agent receives one action from its imitation policy and one from its RL actor, and centralized critic ensembles choose, per agent, which to execute. The same choice is made inside the TD target, so the critics learn the value of the better of the two policies rather than of the RL actors alone. RL thus improves on the R2BC policy where it is wrong while still following it where it is right, so the demonstrator's ceiling of (i) is not inherited. For (ii), the critics' preference also sets, state by state, how strongly each policy is pulled toward its imitation action (Fig.~\ref{fig:overview}, Learn), so the actors converge to the policy the critics execute and can be deployed alone, without the critics or the imitation policy.

The contributions of this work are:
\textbf{(i)}~We formulate \textbf{IBMARL}, a generalization of IBRL to cooperative Dec-POMDPs in which the imitation-learned (IL) policy is a set of $N$ decentralized R2BC policies and a centralized critic ensemble arbitrates among per-agent combinations of IL and RL proposals.
\textbf{(ii)}~We introduce a \textbf{gated imitation penalty} that pulls each RL policy toward the IL proposal in proportion to the critics' preference for it, so that the actors alone match the centralized policy and execution is decentralized.
\textbf{(iii)}~We evaluate IBMARL on four cooperative tasks in VMAS~\cite{bettini2022vmas} using synthetic demonstrators. Here, IBMARL surpasses the IL policies $2.6$--$4.2\times$ faster than demonstration-based baselines on Navigation and $2.8\times$ faster on the physically coupled Buzzwire task, is the only method to surpass the IL policies on Balance and Transport, and exceeds the performance of the demonstrator on Buzzwire and Balance.
\textbf{(iv)}~We evaluate IBMARL with a real human operator using R2BC's round-robin interface on Buzzwire, where IBMARL surpasses the human-trained IL policy more than twice as fast as the best baseline, suggesting that the gains carry over from synthetic to real demonstrators.

\section{Related Work}
\label{sec:related}
\textbf{Single-Agent RL with Demonstrations:} A significant body of prior work combines expert demonstrations with single-agent RL. In the simplest instantiation, offline demonstrations pretrain the starting policy by behavior cloning~\cite{ankile2025residual,ramrakhya2023pirlnav,rajeswaran2018dapg} or, for off-policy methods, populate the replay buffer with the demonstrated transitions~\cite{hester2018dqfd,nair2020awac}. Both remove much of the early random exploration and give the critic high-value transitions from the start, which stabilizes early training and raises early returns over RL alone~\cite{rajeswaran2018dapg,hester2018dqfd,kang2018pofd,liu2022driving}. A second family treats the demonstrations as offline data and adds a surrogate loss that keeps the policy or critic close to the demonstrated actions: DQfD's large-margin term on the Double DQN loss~\cite{hester2018dqfd,vanhasselt2016double}, which surpasses both imitation and RL alone in nearly all of 42 Atari games within a million steps, and similar extensions in DDPGfD~\cite{vecerik2017ddpgfd}, POfD~\cite{kang2018pofd}, NAC~\cite{gao2018nac}, Hybrid RL~\cite{song2023hybridrl} and TD3+BC~\cite{fujimoto2021td3bc}. When the reward is sparse, the demonstrations provide the signal: Nair et al.~\cite{nair2018overcoming} add a behavior-cloning loss to the DDPG policy and, for suboptimal demonstrations, gate it with a \emph{Q-filter} so that it applies only where the critic rates the demonstrated action above the actor's. IBRL~\cite{hu2023ibrl} keeps the imitation policy itself in the loop, letting the critic choose between its proposal and the actor's at every step (Sec.~\ref{sec:ibrl}). We propose IBMARL, the multi-agent generalization of IBRL; its gated imitation penalty is a soft, state-wise Q-filter applied to the IL policy's action on the states the learner visits rather than to stored demonstrations.

\textbf{Multi-Agent RL with Demonstrations:} Compared to the single-agent literature, MARL extensions that utilize demonstrations are few. Alagha et al.~\cite{alagha2023multiagent} let a synthetic expert take control of one agent during interaction so that part of the replay buffer is expert-provided, which improves early team performance; PRIMAL~\cite{sartoretti2019primal} trains decentralized policies by combining RL with imitation of a centralized planner; others reformulate particular MARL algorithms around demonstrations, namely MADDPG from demonstrations~\cite{qiu2023flocking}, skill learning from demonstrations~\cite{venkatesh2024keypoints} and hybrid training~\cite{zhang2024hybrid}; a separate line imitates opponents in competitive games~\cite{bui2024mimicking}. The ``-fD'' regularizers above could in principle be added to QMIX~\cite{rashid2018qmix}, VDN~\cite{sunehag2018vdn}, FACMAC~\cite{peng2021facmac} or LIIR~\cite{du2019liir}, but to our knowledge MADDPG from demonstrations~\cite{qiu2023flocking} is the only work that has done so. By far the most relevant prior work is PegMARL~\cite{yu2025pegmarl}, which also avoids joint demonstrations: single-agent demonstrations, given with \emph{no other agents present}, train a behavioral discriminator and a transition discriminator that shape the sparse team reward online, leaving all teaching of cooperation to the reward. IBMARL differs in three ways. Its demonstrations are given through R2BC in the presence of teammates, so the IL policy already encodes cooperation; the IL policy acts through action proposals, bootstrap targets and a critic-gated pull on the policy rather than through reward shaping or a scheduled cloning penalty~\cite{nair2018overcoming,fujimoto2021td3bc}; and the output is a set of decentralized actors that need neither the centralized critics nor the IL policy at run time.


\section{Background}
\label{sec:background}
IBMARL combines IBRL, which utilizes an imitation policy with RL, with R2BC, which makes a cooperative imitation policy obtainable for multi-agent systems from one human demonstrator.

\subsection{Imitation Bootstrapped RL}

\label{sec:ibrl}
Imitation Bootstrapped Reinforcement Learning (IBRL)~\cite{hu2023ibrl} is a single-agent off-policy actor-critic method that starts from a behavior-cloned \emph{IL policy} $\mu_\psi$, pretrained on a small set of demonstrations $\mathcal{D}$, and keeps it in the loop in two places. When acting, both policies propose an action for the current observation $o$ and the target critic $\bar{Q}_\phi$ picks the better one,
\begin{equation}
\label{eq:ibrl_act}
a = \operatorname*{arg\,max}_{a' \in \{\mu_\psi(o),\, \pi_\theta(o)\}} \bar{Q}_\phi(o, a'),
\end{equation}
so early exploration follows the IL policy wherever the critic has found nothing better. The same choice is made at the next observation inside the TD target,
\begin{equation}
\label{eq:ibrl_target}
y = r + \gamma \max_{a' \in \{\mu_\psi(o'),\, \bar{\pi}_\theta(o')\}} \bar{Q}_\phi(o', a'),
\end{equation}
so the critic estimates the value of the better of the two policies and value propagates from states the IL policy can reach before the policy can. The replay buffer is pre-loaded with $\mathcal{D}$, and the policy $\pi_\theta$ is trained with the ordinary deterministic policy gradient on $Q_\phi$. With 10--50 demonstrations and a poor IL policy, IBRL reaches 85--100\% success on real-robot manipulation tasks. Two features matter for what follows. First, the IL policy only ever supplies an action for the critic to pick or reject; it never enters the actor's loss, so the policy itself is not trained toward it. Second, the propose-and-select step is also used at test time, so that the critic and IL policies are required for deployment.

\subsection{Round-Robin Behavior Cloning}
\label{sec:r2bc}
Round-Robin Behavior Cloning (R2BC)~\cite{mattson2026r2bc} is an online multi-agent imitation learning method that learns a cooperative policy for $N$ agents from one demonstrator who controls a single agent at a time. A session proceeds in rounds. In each round, the demonstrator controls one agent $i$ through a teleoperation interface for one or more episodes while every other agent $j \neq i$ executes its current learned policy $\mu_{\psi_j}$. Every step of the episode is recorded as a joint transition, and the controlled agent's pairs $(o_i, a_i)$ form its cloning set $\mathcal{D}_i$; we also record the rewards observed at collection so that the transitions can later be replayed by RL. Control then passes to the next agent in round-robin order, and after each round every policy $\mu_{\psi_j}$ is updated by behavior cloning, supervised regression from $o_j$ to $a_j$, on $\mathcal{D}_j$. Because the teammates run their current policies while a demonstration is given, the demonstrator reacts to the team as it actually behaves, which is what lets single-agent demonstrations produce cooperative behavior, and the data is collected on the state distribution the current team visits, which limits covariate shift. In the original work, 30--50 demonstrations per agent suffice for high task performance in simulation and on hardware.
R2BC inherits the limits of behavior cloning: its performance is bounded by the demonstrator's, deployment noise can still take the team to states no demonstration covered, and the last increments of performance cost the most demonstrations. IBMARL addresses these with online RL under a known team reward, using IBRL's mechanism to decide, state by state, when the IL policy is still worth following.
\section{Problem Formulation}
\label{sec:formulation}
Following prior work, we model a cooperative team of $N$ agents as a partially observable stochastic game defined by the tuple $\langle \mathcal{S}, \{\mathcal{A}_i\}, \{\Omega_i\}, P, O, \{r_i\}, \gamma, T \rangle$. At step $t$, agent $i$ receives a private observation $o_{i,t} \in \Omega_i$ of the state $s_t \in \mathcal{S}$ and selects a continuous action $a_{i,t} \in \mathcal{A}_i$; the joint action $\act_t = (a_{1,t}, \dots, a_{N,t})$ moves the team to $s_{t+1} \sim P(\cdot \mid s_t, \act_t)$, and each agent receives a reward $r_{i,t}$, discounted by $\gamma$. Generally, these rewards may differ across agents. We write $\obs_t = (o_{1,t}, \dots, o_{N,t})$ for the joint observation. In our work, we study the cooperative case~\cite{oliehoek2016decpomdp}: a decentralized policy $\pi = (\pi_1, \dots, \pi_N)$, $\pi_i : \Omega_i \rightarrow \mathcal{A}_i$, is scored by its expected team return
\begin{equation}
\label{eq:return}
J(\pi) = \mathbb{E}\Big[\tfrac{1}{N} \textstyle\sum_{t=0}^{T-1} \sum_{i=1}^{N} r_{i,t} \;\Big|\; a_{i,t} = \pi_i(o_{i,t})\Big],
\end{equation}
over episodes of at most $T$ steps.

We utilize the outputs of R2BC (Sec.~\ref{sec:r2bc}): The first is the \emph{IL policy} $\teacher = (\mu_{\psi_1}, \dots, \mu_{\psi_N})$, $\mu_{\psi_i} : \Omega_i \rightarrow \mathcal{A}_i$, frozen once the session ends. The second is the \emph{demonstration set} $\mathcal{D}$ of the session's recorded episodes: in each, the demonstrator, a person or a synthetic policy, controls one agent while the others act with the IL policies of that round, and every step is stored as a joint transition $(\obs, \act, \mathbf{r}, \obs')$ with $\mathbf{r} = (r_1, \dots, r_N)$, so each joint action mixes one demonstrator action with actions of the then-current IL policies. The goal is to learn decentralized RL policies $\pi_{\theta} = (\pi_{\theta_1}, \dots, \pi_{\theta_N})$ that surpass the IL policy, $J(\pi_\theta) > J(\teacher)$, in as few environment steps as possible, and that maximize Eq.~\ref{eq:return} within a limited environment interaction budget. Two assumptions define our application setting:
\textbf{(i)}~\emph{A known reward.} The team reward is given, so the demonstrations are not needed to specify the task, only to make it learnable. The clearest motivating cases for IBMARL are environments with sparse rewards or limited reward signals, as we study in our experiments.
\textbf{(ii)}~\emph{Centralized training, decentralized execution.} During learning, critics may condition on $(\obs, \act)$, but at deployment, each agent executes $\pi_{\theta_i}(o_i)$ alone, with no communication, critics or IL policy. This rules out IBRL's test-time arbitration (Sec.~\ref{sec:ibrl}).

\begin{figure*}[t]
\centering
\includegraphics[width=0.88\textwidth]{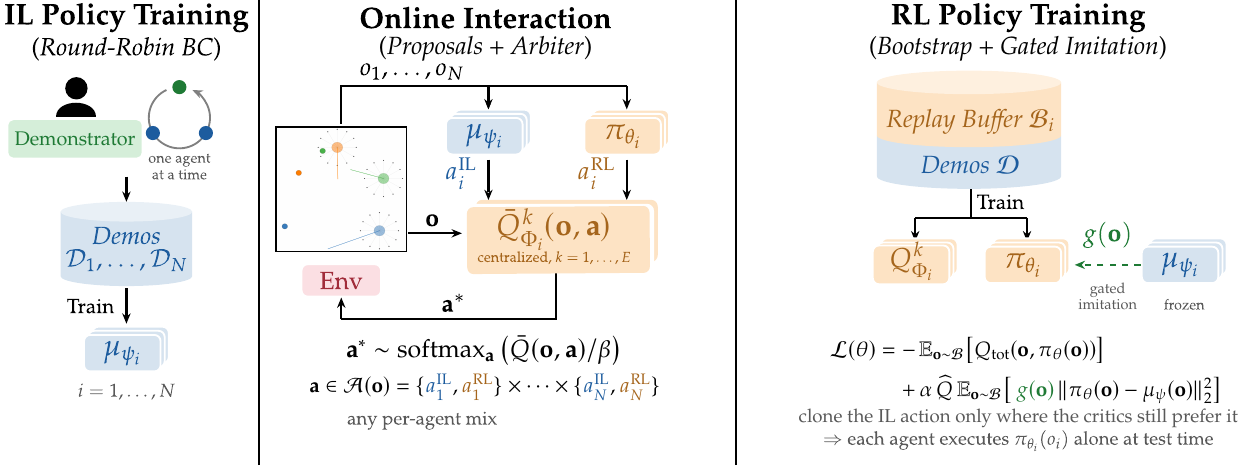}
\caption{IBMARL Pipeline. \normalfont{\textbf{Left:}  R2BC trains the decentralized IL policy $\mu_\psi$ from single-agent demonstrations (Sec.~\ref{sec:r2bc}). \textbf{Middle:} the agents' IL and RL proposals form the $2^N$ joint candidates $\mathcal{A}(\mathbf{o})$, which the target critics score and the arbiter samples (Eqs.~\eqref{eq:candidates}--\eqref{eq:score}). \textbf{Right:} the critics bootstrap through the same arbiter (Eq.~\eqref{eq:target}), and the actor loss adds the gated imitation term (Eqs.~\eqref{eq:actor}--\eqref{eq:gate})}}
\label{fig:pipeline}
\end{figure*}

\section{Imitation Bootstrapped MARL}
\label{sec:method}
IBMARL extends IBRL (Sec.~\ref{sec:ibrl}) to cooperative teams under centralized training with decentralized execution~\cite{lowe2017maddpg} (Fig.~\ref{fig:pipeline}). As in IBRL, the IL policy proposes actions during exploration and for the bootstrap target and pre-populates the replay buffer; unlike IBRL, the choice is made per agent by a critic-scored \emph{arbiter} over $2^N$ joint action candidates, and the same critics gate an imitation term that pulls the RL actors toward the IL actions, so that the actors can be deployed alone. Algorithm~\ref{alg:ibmarl} summarizes the procedure.
 \begin{algorithm}[t]
 \small
\caption{IBMARL}
\label{alg:ibmarl}
\begin{algorithmic}[1]
\Require frozen R2BC IL policies $\mu_{\psi_i}$ and demonstrations $\mathcal{D}$ (joint transitions), $i=1,\dots,N$
\Require actors $\pi_{\theta_i}$; for each agent $i$, an ensemble of $E$ centralized critics $\{Q^k_{\Phi_i}\}_{k=1}^{E}$; target copies $\bar{\pi}_{\theta_i}$, $\bar{Q}^k_{\Phi_i}$
\Require subset size $|K|$, temperature $\beta$, warm-up size $W$, update interval $F$, gradient steps $G$, minibatch size $M$, Polyak rate $\tau$
\State Initialize the replay buffer $\mathcal{B} \gets \mathcal{D}$
\For{each environment step}
  \State Form noised proposals $a^{\rl}_i, a^{\il}_i$ for all $i$ \hfill\eqref{eq:proposals}
  \State Sample $K\subset\{1,\dots,E\}$; score every $\act \in \cand(\obs)$ with $\bar{Q}$ \hfill\eqref{eq:score}
  \State $\act^\ast \sim \operatorname{softmax}\!\big(\bar{Q}(\obs,\cdot)/\beta\big)$; execute $\act^\ast$; store the transition in $\mathcal{B}$
  \If{$F$ steps have elapsed \textbf{and} $|\mathcal{B}| \ge W$}
    \For{$G$ gradient steps}
      \State Sample $M$ transitions; arbitrate noise-free target proposals at $\obs'$ to get ${\act'}^{\ast}$
      \State Update every critic $Q^k_{\Phi_i}$ toward $y_i$ \hfill\eqref{eq:target}
      \State Compute the gate $g(\obs)$; update actors \hfill\eqref{eq:gate},~\eqref{eq:actor}
      \State Polyak-update target networks with rate $\tau$
    \EndFor
  \EndIf
\EndFor
\State \Return decentralized actors $\{\pi_{\theta_i}\}_{i=1}^N$
\end{algorithmic}
\end{algorithm}

\subsection{Proposals and the Arbiter}
Given a team observation $\obs = (o_1, \dots, o_N)$, each agent forms two candidate actions,
\begin{equation}
\label{eq:proposals}
a^{\rl}_i = \pi_{\theta_i}(o_i) + \varepsilon_i, \qquad
a^{\il}_i = \mu_{\psi_i}(o_i) + \varepsilon'_i,
\end{equation}
with i.i.d.\ exploration noise $\varepsilon, \varepsilon' \sim \mathcal{N}(0, \sigma^2 I)$ on both proposals. The arbiter's candidate set is the product of the per-agent choices,
\begin{equation}
\label{eq:candidates}
\cand(\obs) = \{a^{\il}_1, a^{\rl}_1\} \times \dots \times \{a^{\il}_N, a^{\rl}_N\},
\end{equation}
so it may follow the IL policy for one agent and the RL policy for another. Each agent maintains an ensemble of $E$ centralized critics $\{Q^{k}_{\Phi_i}(\obs, \act)\}_{k=1}^{E}$ with target copies $\bar{Q}^{k}_{\Phi_i}$. Following REDQ~\cite{chen2021redq}, the arbiter samples a subset $K \subseteq \{1,\dots,E\}$ of the target critics, and scores each candidate $\act \in \cand(\obs)$ by
\begin{equation}
\label{eq:score}
\bar{Q}(\obs, \act) \;=\; \sum_{i=1}^{N} \, \min_{k \in K} \, \bar{Q}^{k}_{\Phi_i}\!\left(\obs, \act\right).
\end{equation}
The team acts with a joint action sampled from a Boltzmann distribution over the candidates, $\act^\ast \sim \operatorname{softmax}_{\act \in \cand(\obs)} \big(\bar{Q}(\obs, \act)/\beta\big)$ with temperature $\beta$ (evaluation takes the argmax).

\textbf{Per-agent mixing and credit assignment.} Restricting the arbiter to the two pure joint actions, all-IL or all-RL, would reproduce IBRL's binary choice. Allowing mixed candidates does two things. It lets the team keep a reliable IL action for an agent whose RL policy is still poor while another agent, whose policy has improved, acts on its own, so the centralized policy improves one agent at a time rather than all at once. It also gives the critics a form of counterfactual information: the candidates in $\cand(\obs)$ differ from one another in a single agent's action, so the scores of (Eq.~\ref{eq:score}) compare joint actions that isolate each agent's contribution at the current state. The single-agent demonstrations therefore supply a per-agent learning signal that a joint demonstration would not, which is the signal that credit assignment in cooperative MARL lacks~\cite{albrecht2024marl}. The sum in (Eq. \ref{eq:score}) resembles the additive decomposition of VDN~\cite{sunehag2018vdn}, but each term conditions on the full joint observation and action, so no factorization is assumed; what makes the joint maximization tractable with continuous actions is that it ranges over the $2^N$ candidates rather than over the joint action space itself.

\subsection{Learning}
Critics are trained with a TD(0) target that bootstraps through the arbiter: at the next observation $\obs'$, the arbiter is applied to noise-free proposals from the target actors $\bar{\pi}_{\theta}$ and the IL policy, yielding a bootstrap action $\act'^{\ast}$, and every ensemble member regresses to
\begin{equation}
\label{eq:target}
y_i = r_i + \gamma\,(1 - \mathds{1}{\text{[terminal]}})\,
\min_{k \in K} \bar{Q}^{k}_{\Phi_i}(\obs', \act'^{\ast}),
\end{equation}
so members differ only by initialization. This is the multi-agent form of (Eq. \ref{eq:ibrl_target}): the target evaluates the best arbitrated candidate at $\obs'$ rather than the actors' own joint action.

\textbf{gated imitation penalty.} The actors are trained with the deterministic policy gradient through the centralized critics, as in MADDPG~\cite{lowe2017maddpg,lillicrap2016ddpg}, but on the team objective $Q_{\text{tot}}(\obs, \pi_{\theta}(\obs)) = \sum_i Q^{1}_{\Phi_i}(\obs, \pi_{\theta}(\obs))$, in which every actor's current action enters the joint action and one ensemble member per agent suffices. This gradient only climbs the critics locally around the actors' own actions. When the arbiter executes IL actions, their successes raise the critics' value of those actions, but nothing in the gradient moves the actors toward them, so the RL actors lag behind the centralized policy that is actually executed (Sec.~\ref{sec:ablations}). We therefore add an imitation penalty to the policy loss:
\begin{equation}
\label{eq:actor}
\mathcal{L}(\theta) = -\mathbb{E}\big[Q_{\text{tot}}(\obs, \pi_{\theta}(\obs))\big] + \alpha\hat{Q}\,\mathbb{E}\big[g(\obs) \lVert \pi_{\theta}(\obs) - \teacher(\obs) \rVert_2^2\big],
\end{equation}
where the gate $g(\obs)$ is the critics' preference for the IL policy at $\obs$, the Boltzmann (sigmoid) probability of the pure IL joint action over the actor's noise-free joint action under the scoring of (Eq. \ref{eq:score}),
\begin{equation}
\label{eq:gate}
g(\obs) = \sigma\big(\big[\bar{Q}(\obs, \teacher(\obs)) - \bar{Q}(\obs, \pi_{\theta}(\obs))\big]/\beta_g\big),
\end{equation}
with gate temperature $\beta_g$. $\hat{Q}$ is the detached minibatch mean of $\lvert Q_{\text{tot}} \rvert$, so that $\alpha$ is dimensionless and transfers across tasks with different return scales, as in TD3+BC~\cite{fujimoto2021td3bc}. The penalty clones the IL action only where the critics still prefer it and switches itself off, state by state, as the policy overtakes the IL policy. It needs no annealing schedule and acts on the states the learner visits rather than on the demonstration set. It is in this sense a soft version of the Q-filter of Nair et al.~\cite{nair2018overcoming}, with the IL policy standing in for the stored demonstrations, and it recovers the hard filter as $\beta_g \to 0$.

Two properties follow from (Eq. \ref{eq:gate}). First, the term acts only where the policy differs from the IL policy and when the critics prefer the IL action. These are the exact states where the arbiter would override the actor, as elsewhere either the squared distance or the gate vanishes. Second, the gate uses the same target critics and scoring as the arbiter, so the policy is pulled toward the action the team would execute at that state, never toward one the arbiter would reject. The actors are therefore trained to reproduce the actions the arbiter would execute, so after training each agent can run $\pi_{\theta_i}(o_i)$ alone.

\subsection{Data and Implementation}
A single replay buffer, shared by all agents, is pre-loaded with $\mathcal{D}$ and sampled uniformly: online transitions gradually dilute the demonstrations. Each stored transition carries the per-agent rewards $\mathbf{r}$, and agent $i$'s critics regress to the target of (Eq.~\ref{eq:target}) built from its own $r_i$. Updates begin after the buffer holds $W$ transitions. Optionally, as in IBRL, the IL policy’s noised proposal is executed unconditionally during warm-up so that the first arbitrated decisions follow observed successes. Relative to IBRL, the only new hyperparameters are the weight $\alpha$, made scale-free by the normalization of (Eq. \ref{eq:actor}), and the gate temperature $\beta_g$.

\begin{figure*}[t]
\centering
\includegraphics[width=0.9\textwidth]{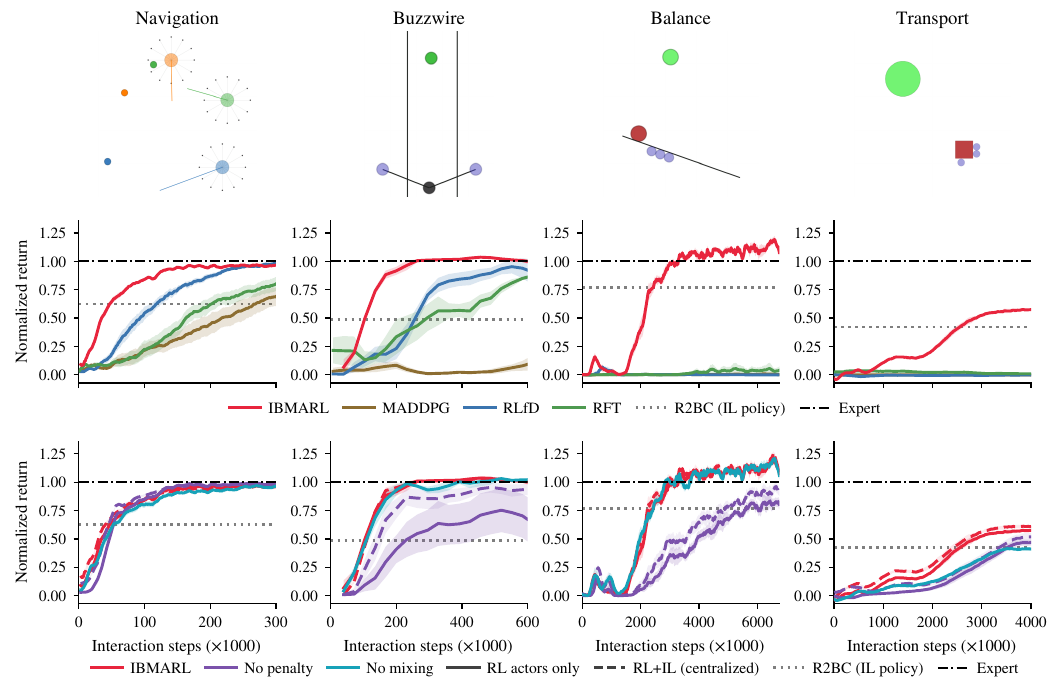}
\caption{Synthetic demonstrations. \normalfont{\textbf{Top:} the four VMAS tasks (Navigation adapted from~\cite{mattson2026r2bc}). \textbf{Middle:} normalized evaluation return of the RL actors (mean $\pm$ s.e.m., 5 seeds); dotted line: frozen R2BC IL policy; dash-dotted line: the expert demonstrator. \textbf{Bottom:} ablations; solid curves are the RL actors alone, dashed the centralized (arbitrated) policy.}}
\label{fig:main}
\end{figure*}

\section{Experiments with Synthetic Demonstrations}
\label{sec:results}

\subsection{Setup}
\label{sec:setup}

\textbf{Tasks.} We evaluate on the \textit{Navigation}, \textit{Buzzwire}, \textit{Balance} and \textit{Transport} tasks of the Vectorized Multi-Agent Simulator (VMAS)~\cite{bettini2022vmas} (Fig.~\ref{fig:main}, top).
In \textit{Navigation}, $N{=}3$ agents and their assigned goals spawn at random in a bounded 2-D workspace. Each agent observes its own state, its goal's relative position, and a LiDAR scan of its teammates.
In \textit{Buzzwire}, $N{=}2$ agents are connected to a ball and must thread it through a narrow corridor to a goal without touching the walls. Each agent observes its own position, velocity and relative position to the goal location.
In \textit{Balance}, $N{=}3$ agents support a rigid beam to lift a spherical package resting on top to a goal without the package or beam touching the floor. Each agent observes its own position and velocity, its position relative to the package and to the beam, the package's position relative to the goal, the package's velocity, and the beam's velocity, angular velocity and rotation.
In \textit{Transport}, $N{=}3$ agents must collaborate to push a package onto a goal. Each agent observes its own state and the package's state relative to the goal and to itself.
In all four, agents act through continuous velocity commands over a fixed horizon $T$.

\textbf{Rewards.} The native VMAS rewards of Navigation, Balance, and Buzzwire are dense, which poorly reflects the difficulty of real world reward signals, which are often goal-conditioned and sparse. We replace these environment rewards with sparse variants, $r_{i,t} = \tilde{r}_{i,t}$ if $d_{i,t} < \delta$ and $-1$ otherwise, where $d_{i,t}$ is agent $i$'s distance to its goal (Navigation) or the ball's distance to the goal, shared by both agents (Buzzwire), and $\tilde{r}_{i,t}$ is the native VMAS reward. Buzzwire episodes run the full horizon and Navigation episodes end early only once every agent is on its goal, so an agent's return counts the steps spent outside the goal region and is bounded below by $-T$. Balance uses a binary success reward: the team receives 1 on the step the package reaches the goal, which ends the episode, and 0 otherwise; the episode also ends if the package or the beam touches the floor, so the return is the success indicator. Transport keeps its native VMAS reward, the per-step decrease of the package's distance to the goal, which is already a difficult exploration signal: the agents must explore an open environment and make contact with the package before any reward is obtained. On both, the difficulty is coordination: one agent cannot lift the beam, and one agent moves the package too slowly to reach the goal within the horizon.

\textbf{Experts, IL policies and demonstrations.} On each task the IL policy is the decentralized policy of one R2BC training session (one small MLP per agent), trained from demonstrations given through R2BC's round-robin protocol by a synthetic demonstrator, the task's ground-truth \emph{expert}: a PPO policy trained with the dense reward on Navigation, a MAPPO~\cite{yu2022mappo} policy on Balance, and scripted controllers that hold each agent at a fixed lateral offset while driving toward the goal on Buzzwire and line the agents up behind the package on Transport. The experts themselves are far from perfect: the Transport expert delivers the package in 28\% of its episodes and the Balance expert lifts it to the goal in 37\%. Their returns are listed in the first two rows of Tab.~\ref{tab:main}. All demonstration-based methods consume the same transitions and IL policy.

\textbf{Implementation.} Actors and critics are per-agent MLPs with two hidden layers of 256 units with $\tanh$ activations. IBMARL uses $E=3$ critics with $|K|=2$, arbiter temperature $\beta=0.05$ and imitation weight of $\alpha=0.4$. All networks are trained with Adam at a learning rate of $10^{-4}$, batch size 1024, $\gamma=0.99$, Polyak $\tau=10^{-3}$, exploration noise $\sigma=0.1$ and gradient norm clipping at $1.0$. On Buzzwire the IL policy is executed unconditionally during warm-up.
\textbf{Baselines} share IBMARL's architectures, DDPG-style updates \cite{lillicrap2016ddpg}, exploration noise and hyperparameters, with a single critic per agent, so that \emph{how} demonstrations are used is the only variable. \textbf{MADDPG}~\cite{lowe2017maddpg} is the demonstration-free reference. \textbf{RLfD}~\cite{vecerik2017ddpgfd} keeps the demonstrations in a separate replay buffer from which each minibatch draws a fraction annealed linearly from 50\% to 0\% over training (over the first two thirds on Balance). \textbf{RFT} adds to the policy loss a behavior-cloning regularizer on demonstration states, $\lambda_{BC}\,\lambda(t)\,\lVert \pi_{\theta}(o^{\mathcal{D}}) - a^{\mathcal{D}} \rVert_2^2$~\cite{rajeswaran2018dapg,nair2018overcoming}, with $\lambda_{BC} = 0.1$ and $\lambda(t)$ annealed from 1 to 0 over the first half of training (the whole run on Transport, the first 27\% on Balance): the pull of (Eq. \ref{eq:actor}) without the gate, on demonstration states and a fixed schedule.


\textbf{Evaluation Protocol.} Policies are evaluated periodically without exploration noise on a dedicated environment. Every method, IBMARL included, is evaluated with its RL actors alone, the decentralized policy each agent would deploy; for IBMARL we also report the \emph{centralized} policy, greedy arbitration between the IL and RL proposals, which needs the centralized critics and the IL policy and is therefore a training-time policy, not a deployable one. Returns are smoothed with a centered window of evaluations; intervals are $\pm 1$ s.e.m.\ over 5 seeds, finals average each seed's last 10 evaluations, and steps to a reference level are the first evaluation whose smoothed return exceeds it, median over seeds, with the number of seeds that ever reach it where fewer than five do. To compare tasks on one axis, Fig.~\ref{fig:main} reports the \emph{normalized return} $(R - R_{\text{floor}})/(R_{\text{expert}} - R_{\text{floor}})$, where the floor is the worst attainable return, $-T$ on Navigation and Buzzwire and $0$ on Balance and Transport, so that $0$ is the floor and $1$ is the expert. Per-seed results and Fig.~\ref{fig:main} in raw return units are given in Appendix~C.

\begin{table*}[t]
\caption{Synthetic demonstrations: environment steps until the RL actors surpass the IL policy (median over 5 seeds; a fraction gives the seeds that ever surpass it) and final return (mean $\pm$ s.e.m.\ over 5 seeds of each seed's last 10 evaluations).}
\label{tab:main}
\small
\setlength{\tabcolsep}{5pt}
\begin{tabular}{l rr rr rr rr}
\toprule
 & \multicolumn{2}{c}{Navigation} & \multicolumn{2}{c}{Buzzwire} & \multicolumn{2}{c}{Balance} & \multicolumn{2}{c}{Transport} \\
\cmidrule(lr){2-3}\cmidrule(lr){4-5}\cmidrule(lr){6-7}\cmidrule(lr){8-9}
Method & steps & final & steps & final & steps & final & steps & final \\
\midrule
R2BC (IL policy) & -- & $-48.7 \pm 1.2$ & -- & $-137.5 \pm 1.8$ & -- & $0.287 \pm 0.048$ & -- & $15.7 \pm 1.0$ \\
Expert (demonstrator) & -- & $-17.9 \pm 0.4$ & -- & $-71.2 \pm 1.4$ & -- & $0.372 \pm 0.009$ & -- & $37.2 \pm 0.4$ \\
\midrule
MADDPG & 3/5 & $-43.9 \pm 6.7$ & never & $-194.6 \pm 2.5$ & never & $0.001$ & never & $-0.05 \pm 0.01$ \\
RLfD   & 122k & $\mathbf{-19.6 \pm 0.5}$ & 296k & $-85.9 \pm 8.1$ & never & $0.000$ & never & $-0.06 \pm 0.10$ \\
RFT    & 198k & $-35.0 \pm 4.9$ & 296k & $-111.2 \pm 4.9$ & never & $0.014 \pm 0.014$ & never & $0.35 \pm 0.09$ \\
IBMARL & \textbf{47k} & $-20.9 \pm 0.8$ & \textbf{104k} & $\mathbf{-68.9 \pm 0.7}$ & \textbf{2.33M} & $\mathbf{0.423 \pm 0.017}$ & \textbf{2.63M} & $\mathbf{21.3 \pm 0.3}$ \\
IBMARL (centralized) & -- & $-21.4 \pm 0.9$ & -- & $-68.8 \pm 1.2$ & -- & $0.425 \pm 0.012$ & -- & $22.7 \pm 0.6$ \\
\bottomrule
\end{tabular}
\end{table*}

\subsection{Comparison with Baselines}
\label{sec:baselines}
\textbf{Navigation.} IBMARL's actors surpass the IL policy after a median of 47k environment steps---$2.6\times$ fewer than RLfD (122k) and $4.2\times$ fewer than RFT (198k), while MADDPG reaches it in only 3 of 5 seeds (Fig.~\ref{fig:main}, Tab.~\ref{tab:main}). This is the easiest of the four tasks: the goal regions are large enough that undirected exploration eventually finds them, which is why demonstration-free MADDPG makes steady progress and why, given the full budget, replay-based RLfD ($-19.6 \pm 0.5$) ends slightly above IBMARL ($-20.9 \pm 0.8$), both within $0.04$ of the expert in normalized return. RFT ($-35.0 \pm 4.9$) and MADDPG ($-43.9 \pm 6.7$) remain well below. The demonstration-based methods therefore separate on speed rather than on the final level: IBMARL reaches the IL level in a sixth of its budget because the arbiter can execute IL proposals from the first step, so the early replay buffer already holds goal-reaching transitions, whereas RLfD must first improve its actors through the critic before its own rollouts reach the goals.

\textbf{Buzzwire.} This task is considerably harder: the ball must be carried through the corridor without contact, so random joint exploration essentially never observes a reward, and MADDPG never reaches the goal in any seed ($-194.6 \pm 2.5$). IBMARL's actors surpass the IL policy after 104k steps, $2.8\times$ fewer than RLfD and RFT (296k each), and finish at $-68.9 \pm 0.7$ against $-85.9 \pm 8.1$ for RLfD and $-111.2 \pm 4.9$ for RFT. The curves in Fig.~\ref{fig:main} show where the difference comes from: the baselines stay at the floor for the first 150--200k steps, the time it takes the critics to propagate value from the demonstration transitions back to the start of the corridor and the actors to follow it, whereas IBMARL executes the IL proposals during warm-up and so observes successes from the beginning. IBMARL is also the first to reach the expert's level, after 264k steps in every seed; RLfD reaches it after 552k steps in 4 of 5 seeds and RFT in 1 of 5, and both have a wide spread across seeds where IBMARL's is tight.

\textbf{Balance.} None of the baselines learns the task within $6.75{\times}10^{6}$ steps: RLfD and MADDPG stay at the floor and RFT barely above it ($0.014 \pm 0.014$). The reward is +1 upon success (zero elsewhere), and an episode ends the moment the package or the beam touches the floor, so a team that does not already know how to lift together drops the package within a few steps and never sees a reward; the demonstration transitions give the critics some positive targets, but the actors cannot reproduce the coordinated lift from them. IBMARL's actors instead surpass the IL policy ($0.287$) after $2.33$M steps, reach the expert's level ($0.372$) after $3.19$M steps in every seed, and end at $0.423 \pm 0.017$, a success rate of 42\% and a normalized return of $1.14$. The IL policy succeeds in 29\% of its episodes, so the arbiter's proposals keep producing successful trajectories on the states the learner actually visits, and the critics learn which of them to keep. The demonstrator is suboptimal but is enough to make the task learnable by IBMARL.

\textbf{Transport.} Within $4{\times}10^{6}$ steps none of the baselines improves beyond its initial performance: the agents must first find the package in an open workspace and push it in a consistent direction before any reward arrives, and one agent alone moves it too slowly to reach the goal, so no baseline learns to push together. IBMARL's actors surpass the IL policy ($15.7$) after $2.63{\times}10^{6}$ steps and end at $21.3 \pm 0.3$ (centralized policy $22.7 \pm 0.6$). Learning is slow because the IL policy itself is weak: a normalized return of $0.42$ against an expert that delivers in only 28\% of its episodes. The arbiter's proposals are often wrong for some agents, but progress comes from the critics gradually learning which agents' IL actions to keep and which actors have surpassed them. Transport is the only task with a large remaining gap to the expert (normalized return $0.57$).

\textbf{Summary.} IBMARL is the fastest to the IL level in all cases, the only method that learns Balance and Transport, and the only one that exceeds the expert (Buzzwire, $1.02$; Balance, $1.14$). The one place a baseline ends ahead is Navigation, where RLfD catches up once exploration is plentiful.

\subsection{Ablations}
\label{sec:ablations}
We conduct ablation tests to justify the individual components of IBMARL. Two ablations isolate the two additions IBMARL makes to IBRL: \emph{No mixing} is IBRL's binary all-IL/all-RL arbiter applied to the team, and \emph{No penalty} is IBRL's objective. Together they measure what the multi-agent setting adds. Fig.~\ref{fig:main} (bottom) plots both the RL actors alone (solid) and the centralized, arbitrated policy (dashed), and Tab.~\ref{tab:ablations} gives the final returns.

\begin{table}[t]
\caption{Ablations: final return (mean $\pm$ s.e.m., 5 seeds) of the RL actors and, where it differs, of the centralized policy.}
\label{tab:ablations}
\footnotesize
\setlength{\tabcolsep}{3pt}
\begin{tabular}{l rrrr}
\toprule
 & Navigation & Buzzwire & Balance & Transport \\
\midrule
IBMARL & $-20.9 \pm 0.8$ & $-68.9 \pm 0.7$ & $0.423 \pm 0.017$ & $21.3 \pm 0.3$ \\
\quad centralized & $-21.4 \pm 0.9$ & $-68.8 \pm 1.2$ & $0.425 \pm 0.012$ & $22.7 \pm 0.6$ \\
No penalty & $-19.4 \pm 0.5$ & $-111.6 \pm 24.0$ & $0.333 \pm 0.015$ & $17.3 \pm 1.0$ \\
\quad centralized & $-19.8 \pm 0.3$ & $-81.6 \pm 6.2$ & $0.347 \pm 0.012$ & $19.2 \pm 1.4$ \\
No mixing & $-21.5 \pm 0.8$ & $-71.2 \pm 3.0$ & $0.422 \pm 0.022$ & $15.3 \pm 0.6$ \\
\bottomrule
\end{tabular}
\end{table}

\textbf{The gated imitation penalty.} On Navigation the term is neutral: both ablations match IBMARL throughout, the actors never lag the centralized policy, so the gate has nothing to close, and \emph{No penalty} ($-19.4 \pm 0.5$) in fact ends marginally above IBMARL. On the three harder tasks the term is decisive, in two different ways. On Buzzwire it closes the \emph{policy lag} of Sec.~\ref{sec:method}: without it the centralized policy still crosses the IL level and ends at $-81.6 \pm 6.2$, but the RL actors alone cross in only 4 of 5 seeds and end at $-111.6 \pm 24.0$, below RLfD. The arbiter keeps choosing the IL action in the corridor, the critics keep rewarding it, and nothing in the deterministic policy gradient moves the actors toward it, so the deployable policy stays worse than the training-time policy. With the term the policy and centralized curves coincide, and the critics and the IL policy can be discarded at deployment. On Balance and Transport the term improves the centralized policy itself, not only the actors: \emph{No penalty} ends at $0.333 \pm 0.015$ (centralized $0.347 \pm 0.012$) on Balance, below the expert, and crosses the IL level later than IBMARL does (Fig.~\ref{fig:main}); on Transport it ends at $17.3 \pm 1.0$ (centralized $19.2 \pm 1.4$) against $21.3 \pm 0.3$ ($22.7 \pm 0.6$), and the centralized policy stays ahead of the actors throughout. Here the pull toward the IL action on the states the learner visits is what lets the actors reach, and then improve on, the coordinated behavior the arbiter was executing, and better actors in turn give the arbiter better RL proposals to choose from.

\textbf{Per-agent mixing.} Mixing is neutral on Navigation, Buzzwire and Balance and decisive on Transport, where \emph{No mixing} ends at $15.3 \pm 0.6$, the IL policy's own level, against $21.3 \pm 0.3$ for IBMARL. On Buzzwire the result is expected: with two agents, mixing adds only two candidates to the pure pair. Balance shows that three agents do not by themselves make mixing useful. The difference is in what the IL policy knows: on Transport the expert succeeds rarely and the IL policy is often right for some agents and wrong for others at the same state, so the arbiter gains from keeping the useful parts of the IL action while letting improved actors act for the rest; on Balance the IL policy tends to be right or wrong for the whole team at once, since all three agents must lift together, and the pure candidates suffice.

\section{Experiments with Human Demonstrations}
\label{sec:human}
The synthetic demonstrators above are consistent and never tire, and the Buzzwire and Transport controllers, in particular, encode a cooperative strategy that a person may not reproduce exactly. R2BC exists to make human demonstrations for teams practical, so we repeat the Buzzwire experiment with demonstrations given by a real human. Using Buzzwire allows us to test human demonstrations for a physically coupled task, where the demonstrator must keep the ball off the walls while a partner they do not control pulls on it, requiring demonstrations that depend on the teammate's behavior.

\begin{figure}[t]
\centering
\includegraphics[width=0.6\columnwidth]{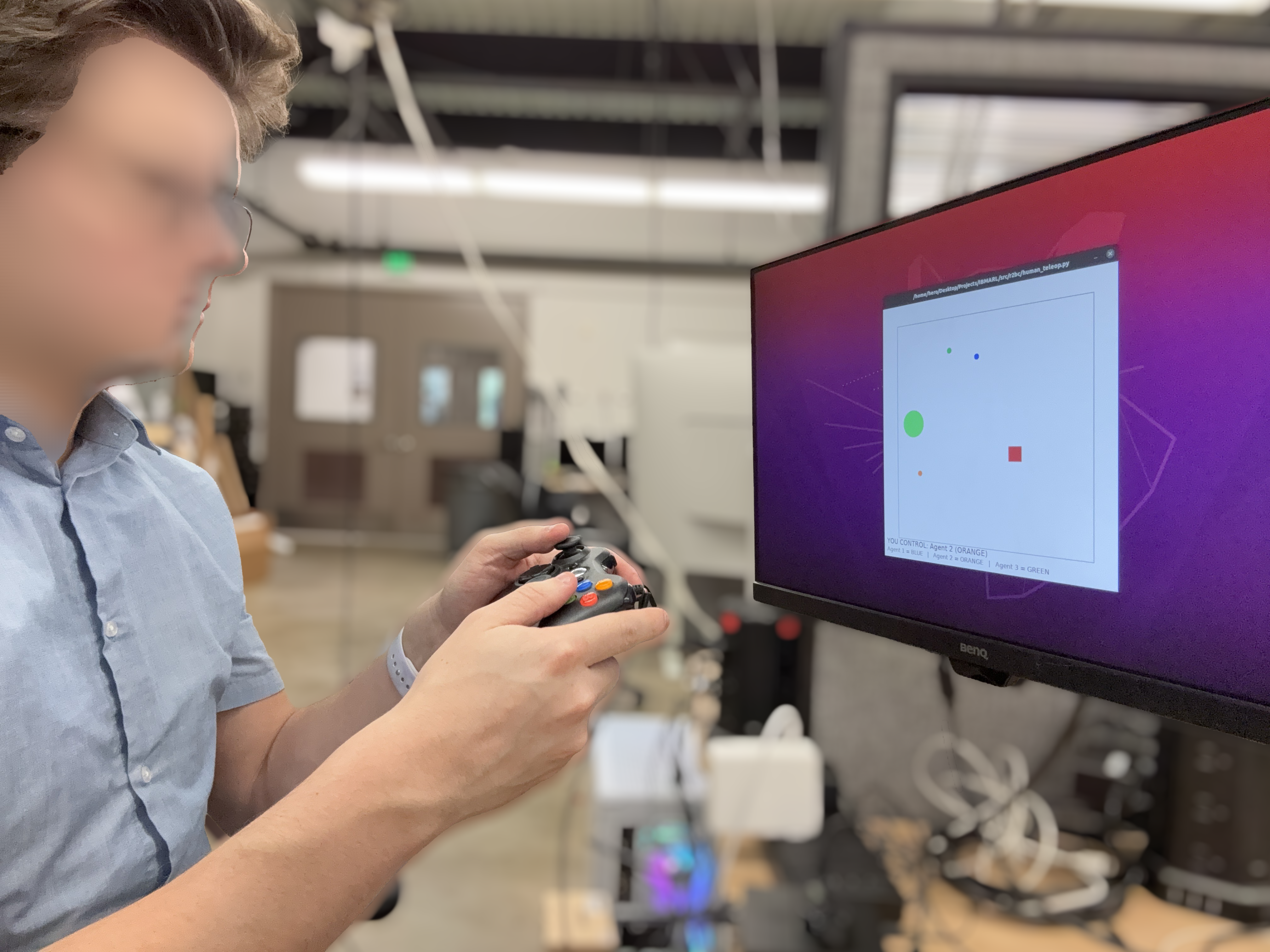}
\caption{Human demonstrations are given through R2BC's teleoperation interface: \normalfont{the demonstrator drives one agent with a gamepad while the others execute their current policies, and control cycles through the team.}}
\label{fig:teleop}
\end{figure}

\textbf{Setup.} A single demonstrator (one of the authors) gives demonstrations through R2BC's teleoperation interface (Fig.~\ref{fig:teleop}), which renders the environment, indicates which agent is currently under the demonstrator's control, and runs the remaining agent on its current R2BC policy. The demonstrator commands the controlled agent's velocity with the analog stick of an Xbox gamepad, and control passes to the other agent in round-robin order, as in R2BC; between rounds the R2BC policies are updated on the data collected so far, so later demonstrations are given alongside a progressively better teammate. The budget is 24 episodes (4{,}800 transitions), twice the scripted budget, of which only 13 reach the goal. The resulting frozen IL policy attains $-143.4 \pm 0.8$, slightly below the expert-trained IL policy's $-137.5$ (normalized $0.44$ against $0.49$). All demonstration-based methods consume the same human transitions and the same human-trained IL policy; the baselines, hyperparameters and protocol are unchanged from Sec.~\ref{sec:setup}.

\begin{figure}[t]
\centering
\includegraphics[width=\columnwidth]{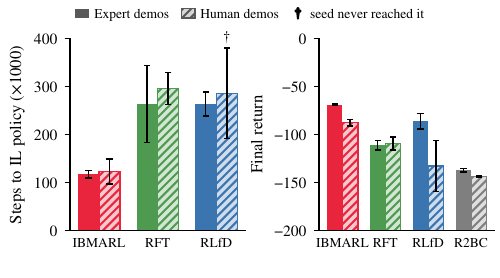}
\caption{Buzzwire with expert and human demonstrations: \normalfont{environment steps until the RL actors surpass the IL policy (left) and final return (right); mean $\pm$ s.e.m.\ over 5 seeds. $\dagger$: one RLfD seed never surpasses the IL policy. R2BC is the frozen IL policy trained from each demonstration set.}}
\label{fig:human}
\end{figure}

\textbf{Results.} The ordering of methods is the same as with synthetic demonstrations (Fig.~\ref{fig:human}; here steps are means over seeds to match the figure). IBMARL's actors surpass the human IL policy after $123\text{k} \pm 26\text{k}$ environment steps in every seed, barely later than with synthetic demonstrations ($117\text{k} \pm 8\text{k}$), against $296\text{k} \pm 34\text{k}$ for RFT and $286\text{k} \pm 94\text{k}$ for RLfD, which fails to surpass it in one seed; IBMARL is thus $2.3$--$2.4\times$ faster than the either baseline. At the end of training IBMARL reaches $-87.7 \pm 3.2$, ahead of RFT ($-109.4 \pm 6.8$) and RLfD ($-132.9 \pm 26.6$), and its seeds agree closely, whereas RLfD's spread is wide. Human demonstrations cost every method some final return, which tracks the lower quality of the human demonstrator, but they cost the baselines more: RLfD, which replays the inconsistent demonstration transitions, falls from $-85.9$ to $-132.9$, close to the IL policy itself, while IBMARL falls from $-68.9$ to $-87.7$ and keeps its lead in both speed and final return; RFT, which clones the demonstration actions on a fixed schedule, is equally poor with either demonstrator. We attribute the difference to how IBMARL uses the IL policies: inconsistent demonstrations are followed where the critics still rate them above the actors and ignored elsewhere.
\section{Discussion}
\label{sec:discussion}
\textbf{Why IBMARL works.} The demonstration-based baselines use the demonstrations only as a stored dataset. IBMARL stores the same dataset but also keeps a cloned IL policy acting throughout the training loop. On Balance and Transport, this policy is the difference between learning and not learning within our budget. The actors are also not bound by the demonstrator: on Buzzwire and Balance they end up outperforming it, so the demonstrator's ceiling, the central limitation of R2BC (Sec.~\ref{sec:r2bc}), is not inherited.

\textbf{Scaling.} IBMARL uses the learned critics to score all $2^N$ joint action candidates. In practice, this only requires one batched forward pass for $N \le 3$, but this approach may not scale for large teams. Because every agent's choice is binary, future work should explore using a greedy linear sweep over agents or a random subset of joint action candidates to replace a full enumeration.

\textbf{Demonstration and deployment cost.} Demonstrations cost is only 12 to 500 episodes of single-agent teleoperation, compared with hundreds of thousands to millions of steps for RL. IBMARL bootstraps from this small amount of human data to achieve significant speedups in MARL. Importantly, at  deployment the team needs only the per-agent actors: IBMARL requires no communication, no centralized critic and no copy of the IL policy.

\textbf{Limitations and future work.} We highlight three limitations that we believe could lead to exciting avenues for future work. First, all results are in simulation, and the human study uses a single demonstrator on a single task. Future work should evaluate IBMARL on real robots and with several demonstrators of varying skill. Second, we assume the team reward is known. While this is a standard assumption in MARL, we note that a reward learned from human preferences~\cite{christiano2017rlhf} could easily replace the ground-truth reward without changing our method. Importantly, this could let a single person supply both the demonstrations and the reward. Third, the IL policy is frozen once RL begins. In the future, the demonstrator could keep teaching during RL and leverage the learned critics to find where the team still needs help; closing the loop would allow a single demonstrator to iteratively fine-tune the behavior of a team of robots, with the demonstrator's effort being spent where it matters.

\section{Conclusion}
\label{sec:conclusion}
Our results demonstrate that realistic, single-human demonstrations can be used to effectively bootstrap multi-agent RL. Crucially, these demonstrations act as a starting point for MARL, rather than a ceiling. Our results also provide evidence that an imitation policy is more useful as a participant than as a data generator. Rather than simply populating a replay buffer, we find value in using an imitation policy to propose actions to be scored by an ensemble of critics. This proposal framework allows them to decide when to imitate and when to diverge from the demonstrations from any state in the environment. 
We see IBMARL as an important step towards multi-robot teams that one person can effectively teach but  that also continue to improve even after the demonstrations end.

\balance
\bibliographystyle{ACM-Reference-Format}
\bibliography{bibliography}

\end{document}